\documentclass[conference]{IEEEtran}
\IEEEoverridecommandlockouts

\title{Deepfake Detection of Face Images based on a Convolutional Neural Network}

\usepackage{cite}
\usepackage{url}
\usepackage{amsmath,amssymb,amsfonts}
\usepackage{algorithmic}
\usepackage{graphicx}
\usepackage{textcomp}
\usepackage{xcolor}
\def\BibTeX{{\rm B\kern-.05em{\sc i\kern-.025em b}\kern-.08em
    T\kern-.1667em\lower.7ex\hbox{E}\kern-.125emX}}

\usepackage{multirow}
\usepackage{placeins}

\begin{document}
\author{\IEEEauthorblockN{Lukas Kroiß}
\IEEEauthorblockA{
\textit{Faculty of Applied Natural Scienes and Cultural Studies} \\
\textit{Ostbayerische Technische Hochschule Regensburg}\\
Regensburg, Germany \\
lukas.kroiss@st.oth-regensburg.de}
\and
\IEEEauthorblockN{Johannes Reschke}
\IEEEauthorblockA{
\textit{Faculty of Electical Engineering and Information Technology} \\
\textit{Ostbayerische Technische Hochschule Regensburg}\\
Regensburg, Germany \\
johannes.reschke@oth-regensburg.de}
}

\maketitle

\begin{abstract} \newline
Fake News and especially deepfakes (generated, non-real image or video content) have become a serious topic over the last years. With the emergence of machine learning algorithms it is now easier than ever before to generate such fake content, even for private persons. This issue of generated fake images is especially critical in the context of politics and public figures. We want to address this conflict by building a model based on a Convolutions Neural Network in order to detect such generated and fake images showing human portraits. As a basis, we use a pre-trained ResNet-50 model due to its effectiveness in terms of classifying images. We then adopted the base model to our task of classifying a single image as authentic/real or fake by adding an fully connected output layer containing a single neuron indicating the authenticity of an image. We applied fine tuning and transfer learning to develop the model and improve its parameters. For the training process we collected the image data set 'Diverse Face Fake Dataset' containing a wide range of different image manipulation methods and also diversity in terms of faces visible on the images. With our final model we reached the following outstanding performance metrics: precision = 0.98, recall 0.96, F1-Score = 0.97 and an area-under-curve = 0.99.
\end{abstract}


\section{Introduction}

Fake news - one of the most current, most popular, but also most problematic topics worldwide. Fake news have probably existed for thousands of years, but have become increasingly important in recent times. Particularly thanks to the internet, the use and spread of fake news have grown exponentially lately. In particular young people use social media platforms on the internet to inform themselves and keeping up-to-date. This becomes especially critical in the context of politics and public figures. \cite{key:Botha_2020}\cite{key:Watson_2023}

Fake news used to appear only in the form of fake texts, but is no longer limited to this single format: a new scope called deepfake can also be counted as fake news. The term is a combination of ‘deep learning’ and ‘fake’. Deepfake therefore refers to the falsification of content, usually images and videos, using generative machine learning algorithms. Since these algorithms are continuously becoming better and cheaper, the creation and use of such media is steadily increasing also by private persons. Fake news in the form of images and videos containing completely new and non-real content can be created and distributed even faster as before using deepfake technologies. \cite{key:Botha_2020}

While machine learning tools are usually developed to create innovative solutions in the field of robotics, autonomous systems or speech models, they can also be used for creating deepfakes enabling these to become increasingly authentic and real. In images for example, certain characteristics or facial expressions of people can be changed or even entire faces can be replaced by another persons face. By doing this, deepfakes are misleading and dramatically influence current public businesses like elections worldwide. Therefore it is increasingly important, but at the same time more complicated, to reliably detect such deepfake content. \cite{key:Botha_2020}\cite{key:Ayuya_2024}

As machine learning algorithms, especially neural networks are used to generate such images, it appears that similar technologies can be used to recognise deepfakes. To counter the misuse of deepfakes, machine learning-based tools are being developed to reliably detect generated images. However, these technologies also have limitations due to rapidly improving generation tools and limited training data, among other things. Nevertheless, many commercial machine learning tools are very promising in terms of deepfake detection. \cite{key:Ayuya_2024}\cite{key:Almars_2021}

Current research projects focus on this field of deepfake detection providing promising results. Rössler et al. for example introduced an advanced benchmark dataset for detecting manipulated facial images and videos in \cite{key:Rossler_2019}. The dataset supports four manipulation methods like DeepFakes (specific method for target face replacement), Face2Face (expression transfer while maintaining the identity of the person), FaceSwap (transfer of a face region or complete face to a target person) and NeuralTextures (facial re-enactment with machine learning based optimizations). Based on this benchmark, they compared different well known network architectures in terms of classifying deepfakes as authentic/real or fake. The best trained model, a XceptionNet architecture (i.e., depth wise separable convolution across all channels including shortcuts like in ResNet), achieved a binary detection accuracy of up to 99.26$\,$\% \cite{key:Rossler_2019}. However the XceptionNet architecture is often not ideal especially for smaller dataset since they are computationally intensive and there is the risk of overfitting \cite{key:Chollet_2017}.

Unlike the traditional method to use the whole face in deepfake detection tasks, Tolosana et al. in \cite{key:Tolosana_2020} explored a method using partial face clues. This method uses specific facial parts such as eyes, nose, or mouth to distinguish between authentic and fake. For the traditional method, again, a Xception model performed best with an area-under-curve (AUC) score up to 100$\,$\%. In contrary, for the second method, the highest AUC score of 100$\,$\% was achieved as well by a Capsule Network (i.e., dynamic routing to capture spatial hierarchies and relationships in data) based on the eyes region of a face \cite{key:Tolosana_2020}. Nevertheless, these kind of networks are computationally intensive and have a complex model structure \cite{key:Stabinger_2024}.

This paper is intended to make a contribution in this area of classifying deepfake images as authentic or fake in terms of a simple and at the same time accurate machine learning algorithm.

We used the ResNet-50 model \cite{key:He_2015} as a basis for our model. This neural network consists of 50 parameter adoptable convolutional layers paired with residual connections every third layers. The top layer of the network is adopted to fit the input images, while the last fully connected layer is replaced by two own layers specific for this task: one flatten layer and one 1-d fully connected dense layer. The very last layer consists of a single output neuron with a sigmoid activation function. The output thus displays a single probability that an authentic image was presented and therefore serves as a classifier. Based on the Diverse Fake Face Dataset (DFFD) \cite{key:Dang_2019}, an AUC score of 0.9931 and a F1-score of 0.9715 were achieved by the trained model.

\section{Methods} \label{sec:methods}
%


We build our model using the ResNet-50 model \cite{key:He_2015} as baseline and adopted this to our task. For training, we applied transfer learning using the DFFD dataset. These aspects will be explained in detail in the following sections.

\subsection{Model}

ResNet-50 \cite{key:He_2015} is a residual convolutional neural network model which uses an advanced model architecture to address the degradation problem occurring in training deep neural nets. Adding more layers to a neural network to build a deep net, causes the accuracy to saturate and at some point degrading rapidly. This effect is not a result of overfitting but is simply caused by the difficulty of training deep neural nets. To counteract this degradation problem, residual networks use identity shortcuts. \cite{key:He_2015} \cite{key:Potrimba_2024}

\begin{figure}[htbp]
\centerline{\includegraphics[width=4cm]{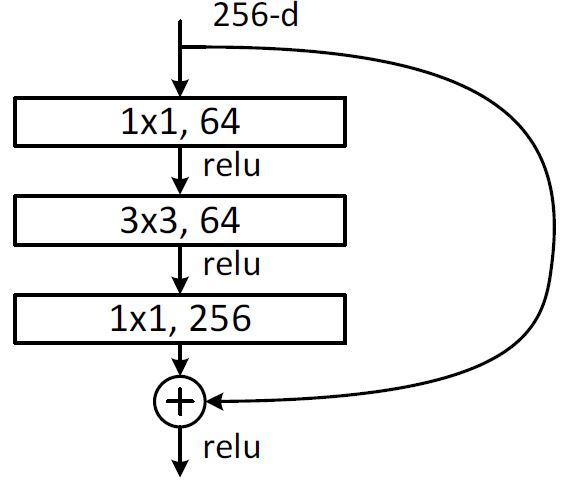}}
\caption{Residual bottleneck building block with shortcut connection \cite{key:He_2015}}
\label{fig:res_block}
\end{figure}

These shortcuts allow a direct flow of information through the net by applying skipping connections to specific parts of the net: the residual block, see Fig. \ref{fig:res_block}. The layers contained in such a block are skipped by the shortcut by adding the input of the building block to its output. This ensures that important information from the input image or from previous layers is forwarded unchanged to deeper layers in the network. The combination of convolutional layers for feature extraction together with shortcut connections allows the ResNet-50 to tackle the degradation and vanishing gradient problem and achieve high accuracy in image classification tasks. \cite{key:He_2015} \cite{key:Potrimba_2024}

In total, the ResNet-50 model consists of 50 parameter trainable layers which are structured in a specific way. Three convolutional layers are always grouped together to form a so-called 'bottleneck' building block, shown in Fig. \ref{fig:res_block}. The 1x1 convolutional layers reduce and afterwards increase, in order to restore, the dimensions, whereas the 3x3 layer in the middle is responsible for feature extraction. Multiple blocks are stacked together to form the whole network. \cite{key:He_2015}

\begin{figure}[htbp]
\centerline{\includegraphics[width=6cm]{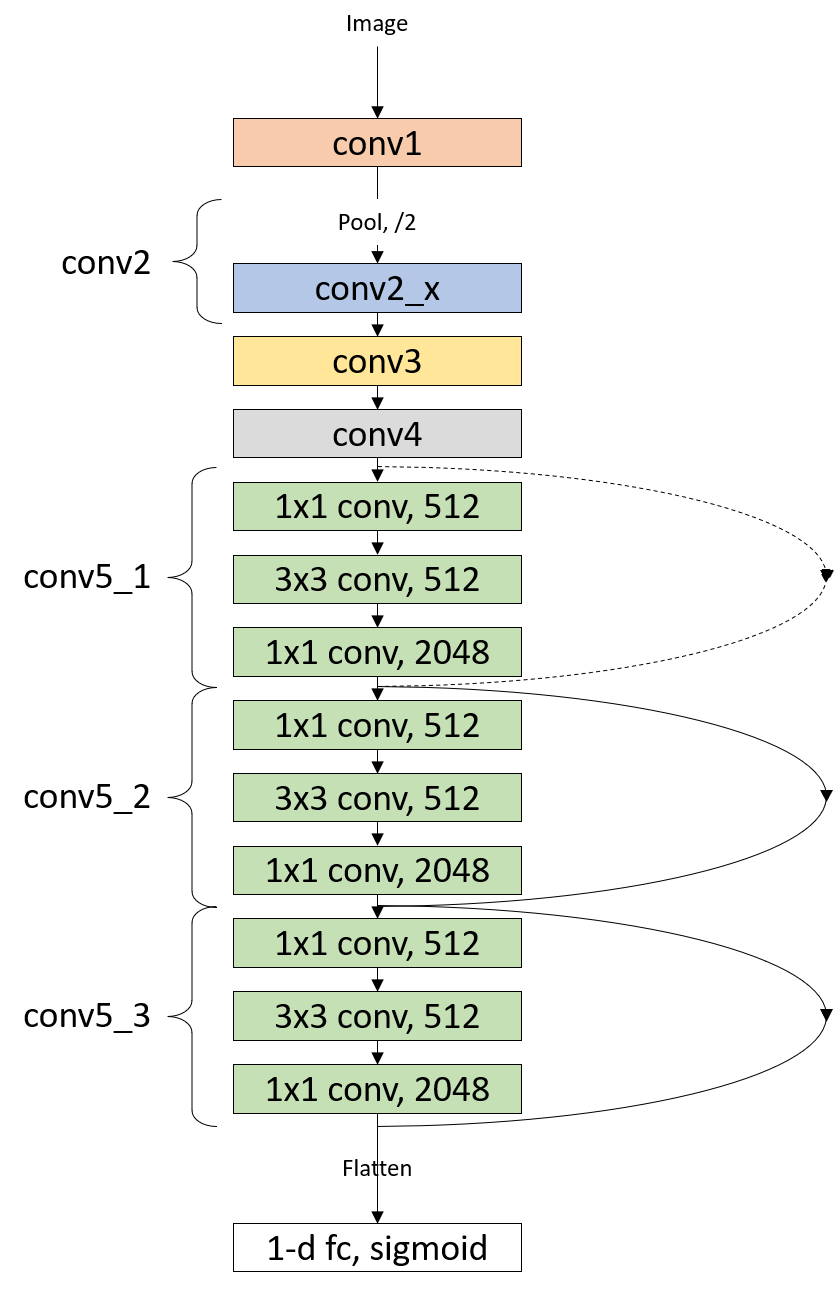}}
\caption{Model architecture of own adopted neural network}
\label{fig:own_model_arch}
\end{figure}

Table \ref{tab:model_arch} shows the architecture for the ResNet-50 model on the left and our adopted model on the right. One building block displayed in Fig. \ref{fig:res_block} is shown in brackets in the table with the number of stacked blocks to the right. The stacked building blocks are grouped by the output size and named from 'conv1' to 'conv5\_x'. Here, the 'x' indicates the incrementing number of the stacked building block in each group. All stacked building blocks together form the whole model. Fig. \ref{fig:own_model_arch} shows the architecture of our adopted model. The stacked blocks of 'conv1' to 'conv4\_x' are still grouped together for the sake of clarity. As an example, the blocks of 'conv5' containing the single convolutional layers together with the shortcut connection skipping every building block are unfolded to display more in details. Between each group of blocks (i.e., each colour in Fig. \ref{fig:own_model_arch}) the image size changes. Therefore the affected shortcuts (dashed) needs to adopt the size of the previous output to match with input size of the following block by applying an additional filter.

\begin{table*}[htbp]
\begin{center}
\caption{Architectures of models, \textbf{Left}: original ResNet-50 model \cite{key:He_2015}, \textbf{Right}: adopted ResNet-50 model}
\label{tab:model_arch}
\newcommand{\ResNetConvB}{
$\left[\begin{array}{c}
1 \times 1, 64 \\
3 \times 3, 64 \\
1 \times 1, 256
\end{array}\right] \times 3$
}

\newcommand{\ResNetConvC}{
$\left[\begin{array}{c}
1 \times 1, 128 \\
3 \times 3, 128 \\
1 \times 1, 512
\end{array}\right] \times 4$
}

\newcommand{\ResNetConvD}{
$\left[\begin{array}{c}
1 \times 1, 256 \\
3 \times 3, 256 \\
1 \times 1, 1024
\end{array}\right] \times 6$
}

\newcommand{\ResNetConvE}{
$\left[\begin{array}{c}
1 \times 1, 512 \\
3 \times 3, 512 \\
1 \times 1, 2048
\end{array}\right] \times 3$
}

\begin{tabular}{l||c|c||c|c}
\hline
      & \multicolumn{2}{c||}{\textbf{ResNet-50}} & \multicolumn{2}{c}{\textbf{Our Model}} \\ \hline
layer name & output size & layers & output size & layers \\ \hline
input & 224 x 224 &       & 299 x 299 &  \\ \hline
conv1 & 112 x 112 & 7 x 7, 64, stride 2 & 150 x 150 & 7 x 7, 64, stride 2 \\ \hline
\multirow{2}[2]{*}{conv2\_x} & \multirow{2}[2]{*}{56 x 56} & 3 x 3 max pool, stride 2 & \multirow{2}[2]{*}{75 x 75} & 3 x 3 max pool, stride 2 \\ \cline{3-3} \cline{5-5}
      &       & \ResNetConvB &       & \ResNetConvB \\  \hline
conv3\_x & 28 x 28 & \ResNetConvC & 38 x 38 & \ResNetConvC \\ \hline
conv4\_x & 14 x 14 & \ResNetConvD & 19 x 19 & \ResNetConvD \\ \hline
conv5\_x & 7 x 7 & \ResNetConvE & 10 x 10 & \ResNetConvE \\ \hline
output   & 1 x 1 & average pool, 1000-d fc, softmax & 1 x 1 & flatten, 1-d fc, sigmoid \\ \hline
\end{tabular}%
\end{center}
\end{table*}

As can be seen in table \ref{tab:model_arch} and Fig. \ref{fig:own_model_arch}, we used the ResNet-50 model as base for our own model and adopted it to our task. Firstly, the size for the input images is increased from 224x224 in the original model to 299x299 in order to keep the original size of the portrait images. This implicitly causes the change of the output size of every layer. Nevertheless, this does not affect the output of the convolutional layers itself, but just the dimensions of its feature maps. 

On the other side, we also changed the output layers to reflect our task. While there were 1000 categories to be classified by the original ResNet-50, our model only has to provide one single output value as prediction. Therefore, we replaced the last two layers in the ResNet-50 (average pooling layer and 1000-d fully connected layer) with two randomly initialised layers: a flatten layer and a 1-d fully connected (dense) layer. We applied a sigmoid activation function to the last layer instead of a softmax classifier, since just one output was required.

According to the sigmoid function in equation \ref{eqn:sigmoid}, the single output neuron reflects a prediction value between 0 and 1 on the authenticity of the input image. We chose sigmoid over softmax activation function for our model due to the  following facts: First, the reduced parameters to be trained and thus the reduced training effort and time \cite{key:Zhenghao_2021}, and second, that sigmoid with a single output neuron is enough for a binary classification task \cite{key:Bengio_2016}.

\begin{equation}
\label{eqn:sigmoid}
\sigma(x) = \frac{1}{1 + e^{-x}}
\end{equation}

After building the model to form the neural net described above, we compiled it with two further parameters in order to prepare it for the training: first, the optimization algorithm and second the loss function. To optimize the training process of our model, we used the Adam optimizer \cite{key:Kingma_2014}. This algorithm optimizes and adopts the learning rate during the training process \cite{key:Bengio_2016} what makes it an effective optimization algorithm \cite{key:Brownlee_2017}. As configuration parameters we used a $learning ~rate = 0.001$, $beta1 = 0.9$ and $beta2 = 0.999$ as suggested by Kingma and Ba in \cite{key:Kingma_2014}. For the parameter $epsilon$ we differ from the authors proposal and used the default setting from the tensorflow library which is $epsilon = 1\times 10^{-7}$. Second, as loss function we used the binary cross entropy (also known as log loss) function since it is commonly used for binary classification problems \cite{key:Bengio_2016}.


\subsection{Dataset} \label{sec:dataset}

The dataset used to train our model is the 'Diverse Fake Face Dataset' (DFFD) \cite{key:Dang_2019}, consisting of real and fake portrait images of human faces. Compared to other datasets, it contains more diversity in terms of age of the people, face size, race and manipulation types. To achieve this, the dataset is a large collection of different datasets for human faces. Although the dataset contains a total of 781,727 samples for real images and 1,872,007 for fake images \cite{key:Dang_2019}, not all of them were used to train our model. Some images were sorted out manually due to a bad quality or not showing an human portrait at all.

In detail, the DFFD consists of eleven different data sets \cite{key:Dang_2019}. For real data, images from the CelebA \cite{key:Liu_2015}, FFHQ \cite{key:karras_2018}, and FaceForensics++ \cite{key:Rossler_2019} datasets were collected. Images from Faceswap \cite{key:Rossler_2019}\cite{key:GitHub_2024}, Deepfake \cite{key:Rossler_2019}\cite{key:GitHub_2024}, Face2Face \cite{key:Rossler_2019}\cite{key:Thies_2020}, and DeepFaceLab \cite{key:Admin_2021} contribute as identity and expression swap fake images. Based on images from CelebA and FFHQ attribute manipulated fake images were generated using the FaceApp \cite{key:faceapp_2019} and the StarGAN \cite{key:yunjey_2018}. Using PG-GAN \cite{key:karras_2017} and StyleGAN \cite{key:karras_2018}, entire fake faces were synthesized based on different input datasets. Images of all these datasets collected in the DFFD \cite{key:Dang_2019} should have a standard images size of 299x299.

Not all data obtained by the dataset mentioned above contains images. The FaceForensics++ dataset consists of 1000 videos. By manipulating these videos in different manners, the three datasets Faceswap, Deepfake and Face2Face were generated which contribute with 3000 fake videos \cite{key:Rossler_2019}. To convert these videos into images, we extracted only one frame per second from every video to reduce the number of similar looking images. In this way, also these datasets containing originally only videos were available as images.

Furthermore images of some datasets contained in the DFFD were not aligned correctly to show a portrait of a human face, whereas images of other datasets were already correctly aligned. To be able to use all the datasets in the same manner and to work best to train our model, the images coming from CelebA, FaceForensics++, Faceswap, Deepfake and Face2Face had to be aligned by us. Using the provided Matlab functions for cropping images and using files containing bounding boxes coordinates for every single image \cite{key:Dang_2019}, we were able to align all affected images in order to reflect the default image size of 299x299. In the appendix, Fig. \ref{fig:alignment} shows an examples of an unaligned image together with the corresponding image after the alignment via Matlab was done. After these alignments, all images were correctly formatted for the following training and testing steps.

Considering all these datasets, a total amount of over 2.6m images were collected, but not all were provided by DFFD. To balance the dataset, Dang et. al. randomly selected 58,703 real  and 240,336 fake images\cite{key:Dang_2019}. For our task, we only used a subset out of this. We did not use the DeepFaceLab dataset since the data was not accessible for free. Anyway, this additional dataset was deemed as not necessary since the remaining datasets contain enough images for our task. Also the Faceswap and StarGan datasets were not used since we deemed the manipulation quality of the images as too bad. We deselected further images of all other sets manually based on the image quality or a wrong face alignment in case the images did not show a portrait. Fig. \ref{fig:unused} in the appendix shows samples of images which were deselected and thus not used. This results in a number of images per dataset as shown in table \ref{tab:datasets_classes_sum}.

\begin{table}[htbp]
\begin{center}
\caption{Number of Images per dataset}
\label{tab:datasets_classes_sum}
\begin{tabular}{l|l|r}
Class & Dataset & Count \\ \hline 
\multirow{3}[2]{*}{Real} & CelebA & 19,923 \\  
      & FaceForensics++ & 32,569 \\ 
      & FFHQ  & 19,999 \\ \cline{1-3}
\multirow{7}[2]{*}{Fake} & Deepfake & 32,454 \\ 
      & Face2Face & 32,682 \\ 
      & Face App & 11,809 \\ 
      & PG GAN 1 & 19,943 \\ 
      & PG GAN 2 & 19,962 \\ 
      & Style GAN CelebA & 20,000 \\ 
      & Style GAN FFHQ & 19,996 \\ \hline 
Sum   &       & 229,337 \\
\end{tabular}%

\end{center}
\end{table}

All images shown in table \ref{tab:datasets_classes_sum} are available and can be used for our model. Talking these images, we initially split the data randomly into 75$\,$\% for training, 15$\,$\% for validation and 10$\,$\% for testing. For those splits too, we applied a deselection of some images. The reason for this is, that images in the FaceForensics++, Deepfake and Face2Face datasets were generated out of videos. This results in having similar images (i.e., frames in the original video) per video. To completely separate the test split from the other two splits and to avoid interferences between these three splits, affected images were deleted from the train and the validation split. In detail, whenever the test split contained an image from the above mentioned datasets, the similar images (i.e., frames) coming from the same video source were deleted from the train and the validation split. The detailed number of images per dataset we used can be seen in table \ref{tab:datasets_classes_used} and the splitting of the dataset in table \ref{tab:datasets_classes}. Fig. \ref{fig:datasets} in the appendix shows example images of different datasets for each class.

To serve as input images for our model, the pixel values for each image were normalized from [0, 255] to [0, 1].

\begin{table}[htbp]
\begin{center}
\caption{Number of Images per dataset used}
\label{tab:datasets_classes_used}
\begin{tabular}{l|l|rrr|r}
Class & Dataset & Train & Val & Test & Sum \\ \hline 
\multirow{3}[2]{*}{Real} & CelebA & 14,953 & 2,984  & 1,986  & 19,923 \\ 
      & FaceForensics & 2,456  & 464   & 1,833  & 4,753 \\ 
      & FFHQ  & 15,005 & 2,990  & 2,004  & 19,999 \\ \cline{1-6}
\multirow{7}[2]{*}{Fake} & Deepfake & 2,301  & 448   & 1,765  & 4,514 \\ 
      & Face2Face & 2,274  & 431   & 1,929  & 4,634 \\ 
      & Face App & 8,841  & 1,783  & 1,185  & 11,809 \\ 
      & PG GAN 1 & 14,890 & 3,003  & 2,050  & 19,943 \\ 
      & PG GAN 2 & 14,939 & 2,993  & 2,030  & 19,962 \\ 
      & Style GAN CelebA & 15,001 & 3,032  & 1,967  & 20,000 \\ 
      & Style GAN FFHQ & 15,039 & 2,962  & 1,995  & 19,996 \\ \cline{1-6}
Sum   &       & 105,699 & 21,090 & 18,744 & 145,533 \\
\end{tabular}%

\end{center}
\end{table}

\begin{table}[htbp]
\begin{center}
\caption{Train-Validation-Test split}
\label{tab:datasets_classes}
\begin{tabular}{l|rrr|r}
Class & Train & Val & Test & Sum \\ \hline 
Real  & 32,414 & 6,438  & 5,823  & 44,675 \\
Fake  & 73,285 & 14,652 & 12,921 & 100,858 \\\hline
Sum   & 105,699 (72.6\%) & 21,090 (14.5\%) & 18,744 (12.9\%) & 145,533 \\
\end{tabular}%

\end{center}
\end{table}

\subsection{Transfer Learning}

After the first step of building the model and compiling it and the second step of collecting and pre-processing all data in previous sections, this chapter shall explain the whole training process of our neural network.

In training our net, we did not start from scratch but with the pre-trained ResNet-50 model. Since the number of data points (i.e., images) is limited, probably there would not have been enough images to train the whole network on our own. Furthermore, training a randomly initialised model is more time consuming than starting with pre-trained parameters. Therefore we performed transfer learning using already trained weights for the the ResNet-50 base model which were pre-trained on the ImageNet data base \cite{key:ImageNet_2024}. In this way, our training process could concentrate on a full training of the last two layers added to the model by us while the rest of the model only had to be adopted to our binary classification problem.

The whole training process is split up into three steps, which are explained in more detail below. For each step, the training and validation data sets as described in the previous section were used in batches with a size of 16. 

\paragraph{Step 1} In the first training step, the whole network was set as trainable. Thus the parameters for all layers from the ResNet-50 base model and also from our own added dense layer were updated. Overall, a total of 23,739,393 were trainable in this step. By monitoring the validation loss, we applied early stopping with a patience of 5 epochs starting from the $10^{th}$ epoch onwards.

\paragraph{Step 2} For the second training step, the number of trainable parameters was reduced. Only layers from ResNet-50 conv\_5 block (see green layers in Fig. \ref{fig:own_model_arch}) and our added dense layer were not locked. Together, 14,656,001 parameters were trainable in this step. During training, we again applied early stopping monitoring the validation loss and using a patience of 5 epochs but this time it was active from the beginning. 

\paragraph{Step 3} In the last training step, an even further decreased number of trainable parameters was set. In addition to the added dense layer, only the conv\_5\_3 block from ResNet-50 (see last three green layers in Fig. \ref{fig:own_model_arch}) were trainable. This results in a total number of trainable parameters of 3,621,377. Like in the second step, also here early stopping with a patience of 5 epochs based on the validation loss was used from beginning onwards. This last step was mainly for training the output layer.

\subsection{Interpretation of Model Prediction}
After the model is built, compiled, and trained, the actual performance of the network is evaluated using the test data set. This set is completely separated from the training and the validation set and can therefore be used as a representative data set to evaluate the performance of our model. An image of the test data set belongs either to the positive class containing fake images or to the negative class containing authentic images. The actual class of an image is displayed using the label $y$. It is defined as $y \in \{real, fake\} = \{0, 1\} = \{n, p\}$. 

Feeding a test image into the model, we receive a single value as output from the model: $\hat{y}_{raw}$. This single value displays the prediction on the authenticity of the input test image. It is written with the index $raw$ to indicate the unprocessed output of the last dense layer through the sigmoid activation function. Thus, this raw prediction is defined as $\hat{y}_{raw} \in [0, 1]$.

In order to evaluate the model performance, this raw prediction has to be mapped to a binary output to enable a comparison with the actual label $y$. The mapping is done via comparing $\hat{y}_{raw}$ with a threshold $th \in [0, 1]$ as described in equation \ref{eqn:th}. In this way, the predicted label is $\hat{y}$ which is defined as $\hat{y} \in \{0, 1\}$ similar to $y$ . Applying this, there are four possibilities: if the test image is in the positive class and predicted as positive, it is counted under $true ~positive ~(tp)$, whereas it is counted as $false ~negative ~(fn)$ if predicted as negative. Vice versa, if the image belongs to the negative class and is predicted as negative, it is counted as $true ~negative ~(tn)$ but if predicted as positive, counted as $false ~positive ~(fp)$. Since the binary prediction $\hat{y}$ made by the neural network is directly dependent on $th$, also the number of occurrences in $tp$, $fp$, $tn$, and $fn$ are directly influenced by the selected threshold.

\begin{equation}
\label{eqn:th}
\hat{y} = 
\begin{cases} 
0, &  \hat{y}_{raw} < th \\
1, & \hat{y}_{raw} \geq th
\end{cases}
\end{equation}

Usually the threshold is set to $th = 0.5$ by default. With respect to the model performance and the dataset balance (ration of occurrences of authentic and fake images), this might not be always the ideal threshold. This means, that by choosing a better threshold, you can influence the performance of your neural network. In detail, $\hat{y}_{raw}$ will not be influenced, but $\hat{y}$ will change according to the selection of $th$. We use the Receiver Operating Characteristic (ROC) curve \cite{key:Fawcett_2006} to determine the ideal threshold for our neural network. A ROC curve is built by iterating the threshold from 0 to 1. For each threshold, the values for $tp$, $fp$, $tn$, and $fn$ are determined. These are taken to calculate the $false ~positive ~rate ~(fpr)$ and the $true ~positive ~rate ~(tpr)$ according to equations \ref{eqn:fpr} and \ref{eqn:tpr} \cite{key:Fawcett_2006}. In the equations, an indication for the total number of images contained in the negative class is given by $N$ and contained in the positive class by $P$. By plotting the $tpr$ over the $fpr$, the ROC curve is created.

\begin{equation}
\label{eqn:fpr}
fpr = \frac{fp}{fp + tn} = \frac{fp}{N}
\end{equation}

\begin{equation}
\label{eqn:tpr}
tpr = \frac{tp}{tp + fn} = \frac{tp}{P}
\end{equation}

\vspace*{10px}
\section{Results}

\begin{figure}[htbp]
\centerline{\includegraphics[width=8cm]{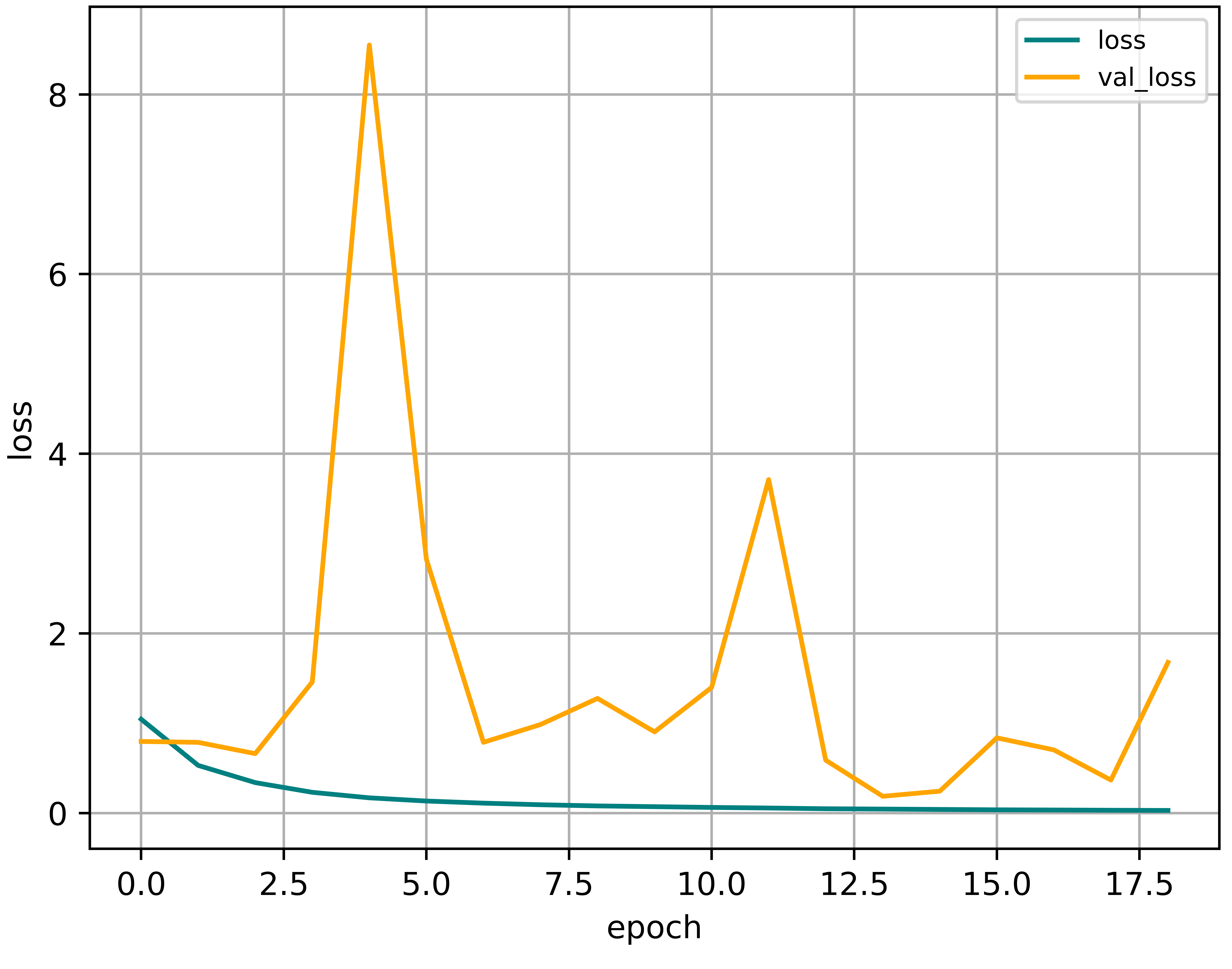}}
\caption{Training Process Step 1}
\label{fig:train_one}
\end{figure}

\begin{figure}[htbp]
\centerline{\includegraphics[width=8cm]{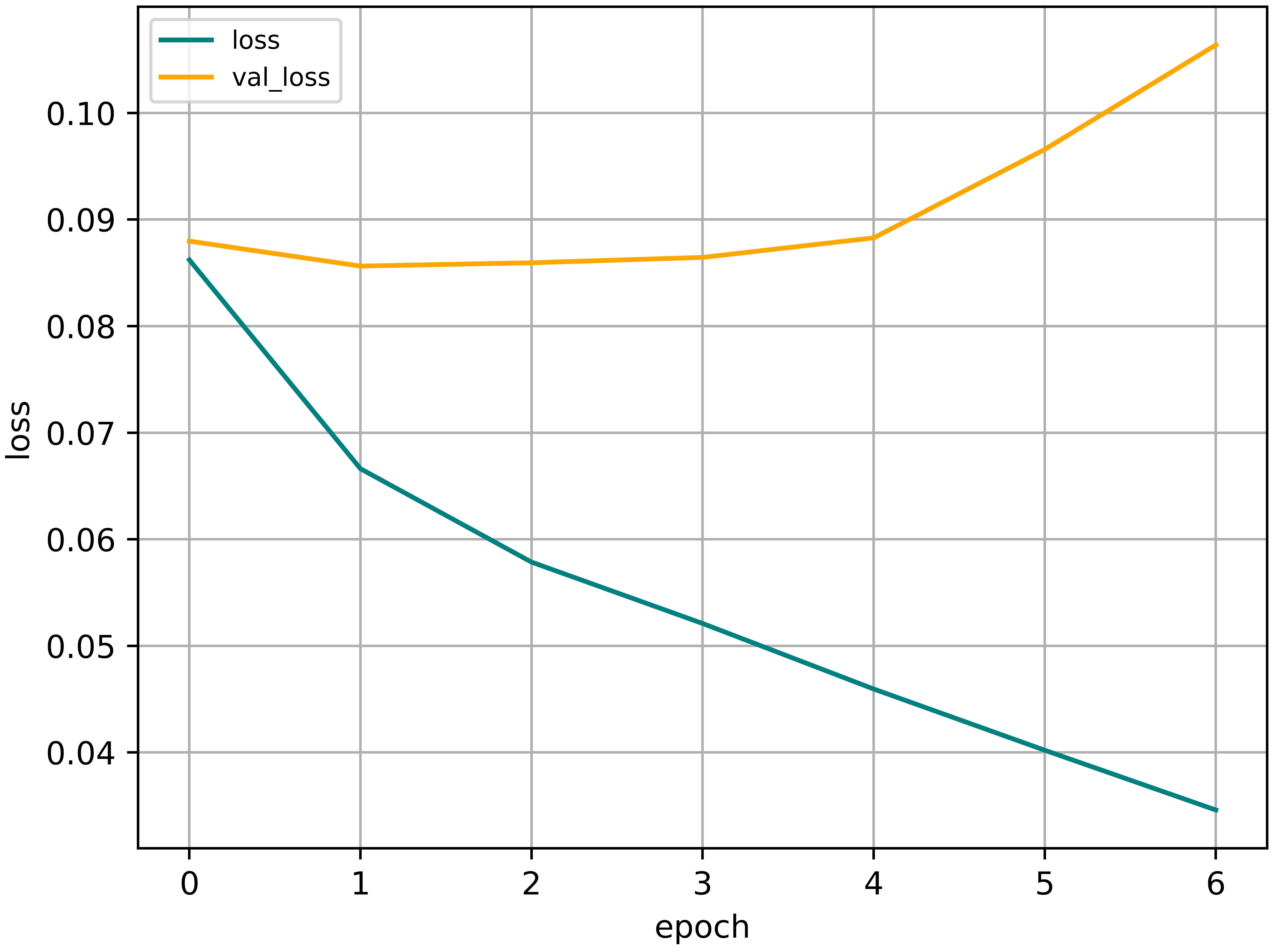}}
\caption{Training Process Step 2}
\label{fig:train_two}
\end{figure}

\begin{figure}[htbp]
\centerline{\includegraphics[width=8cm]{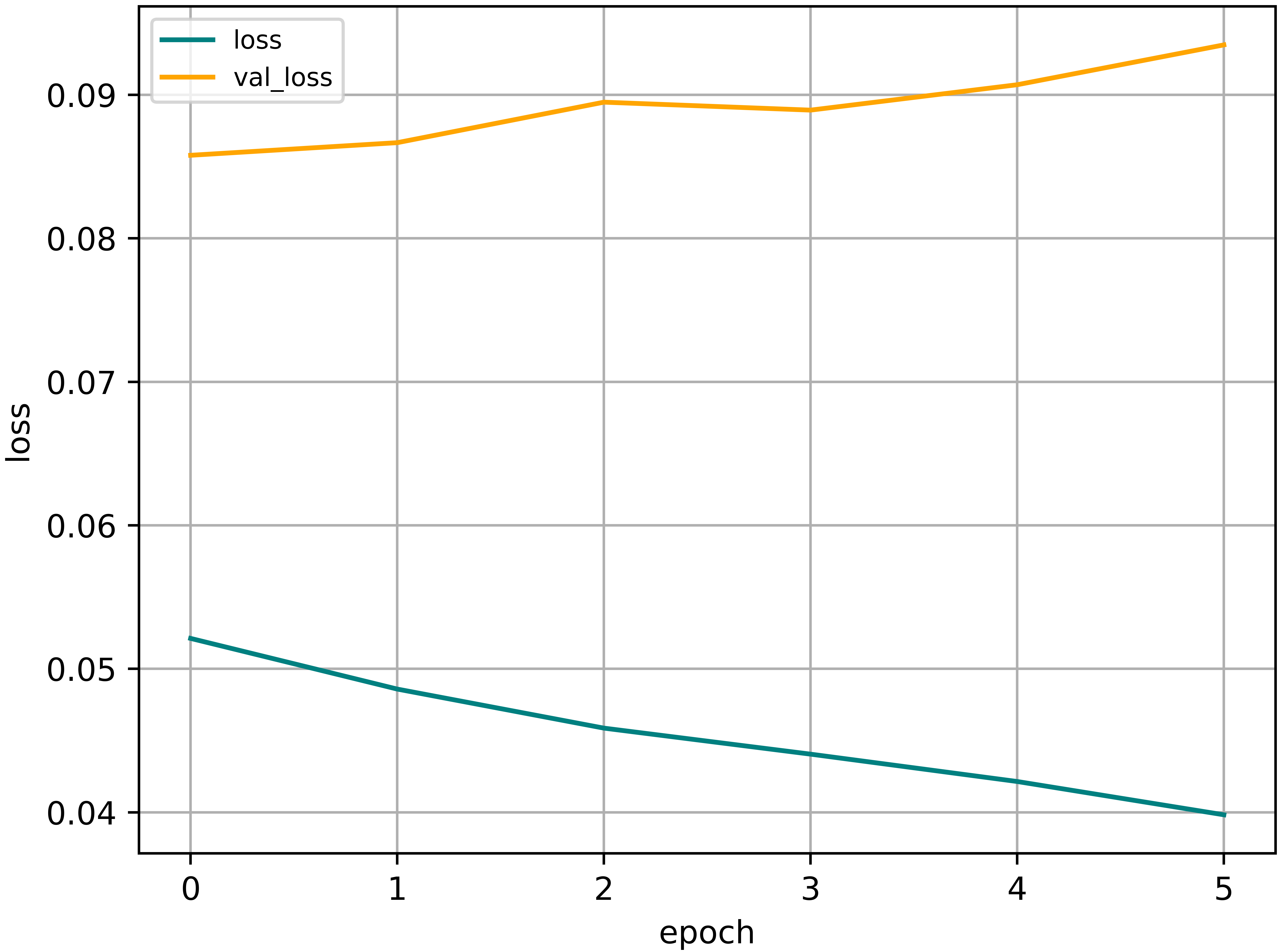}}
\caption{Training Process Step 3}
\label{fig:train_three}
\end{figure}

During the whole training process, the loss based on the training data together with the loss based on the validation data were monitored and evaluated. Fig. \ref{fig:train_one}, \ref{fig:train_two} and \ref{fig:train_three} show these two parameters per epoch for each training step. Since the training process is also split into three steps with different trainable layers, as described in section \ref{sec:methods}, these plots are split as well. For the first step, the training was stopped automatically due to persistent increasing loss measured on the validation data set after the $18^{th}$ epoch. Thus the parameters were effectively trained for 13 epochs. In the second step, it stopped automatically after the $6^{th}$ epoch while in the third step after $5^{th}$ epoch. This means, in the second step, the net is improved for only one epoch, while in the third step, it is not enhanced at all.

\subsection{Receiver Operating Characteristic (ROC)}

\begin{figure}[htbp]
\centerline{\includegraphics[width=8cm]{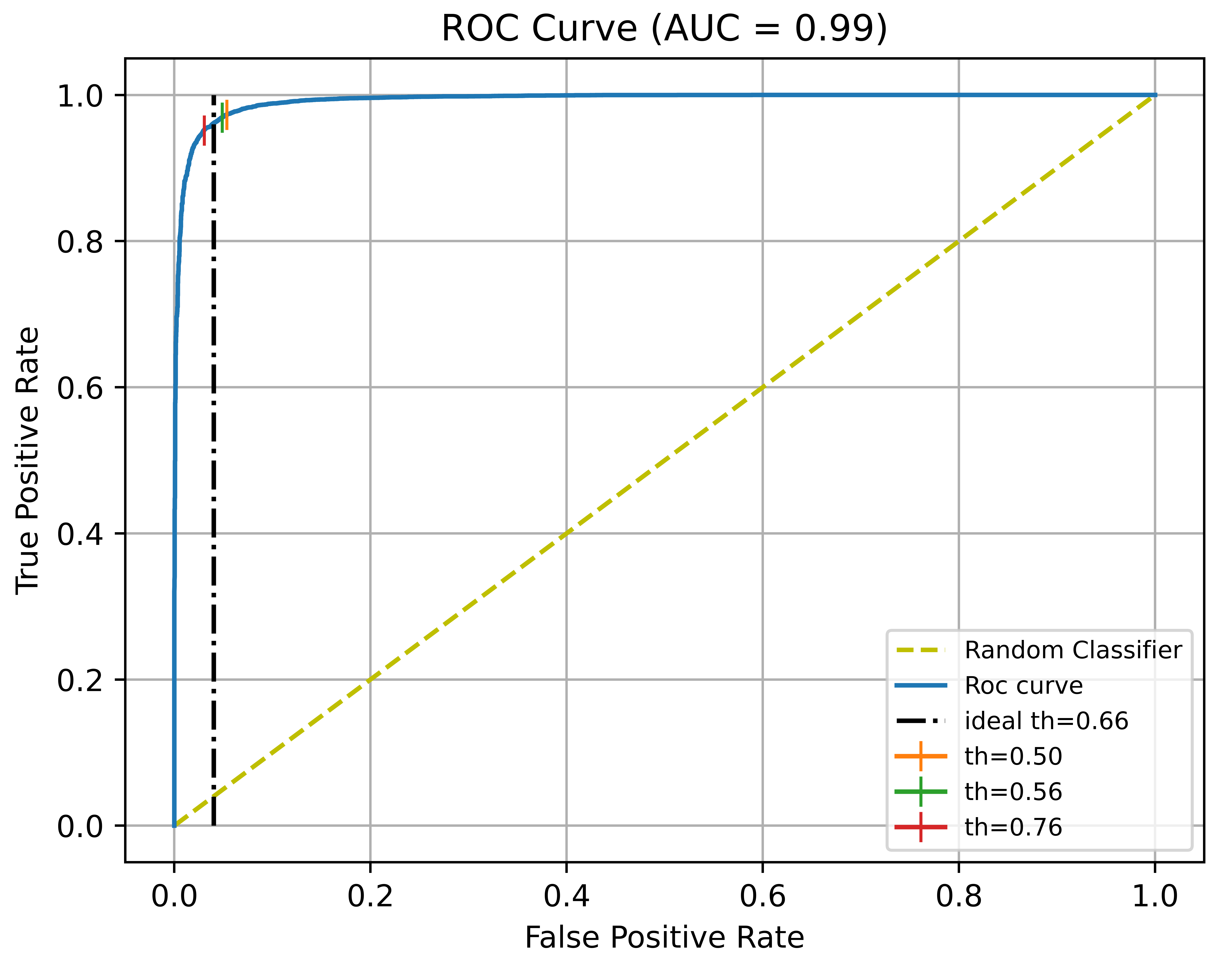}}
\caption{ROC curve with threshold markings. \textbf{Orange:} default threshold = 0.5; \textbf{Black Dashed:} ideal threshold = 0.6587; \textbf{Green:} ideal\_th - 0.1 = 0.56; \textbf{Red:} ideal\_th + 0.1 = 0.76}
\label{fig:roc_curve}
\end{figure}

\begin{figure}[htbp]
\centerline{\includegraphics[width=8cm]{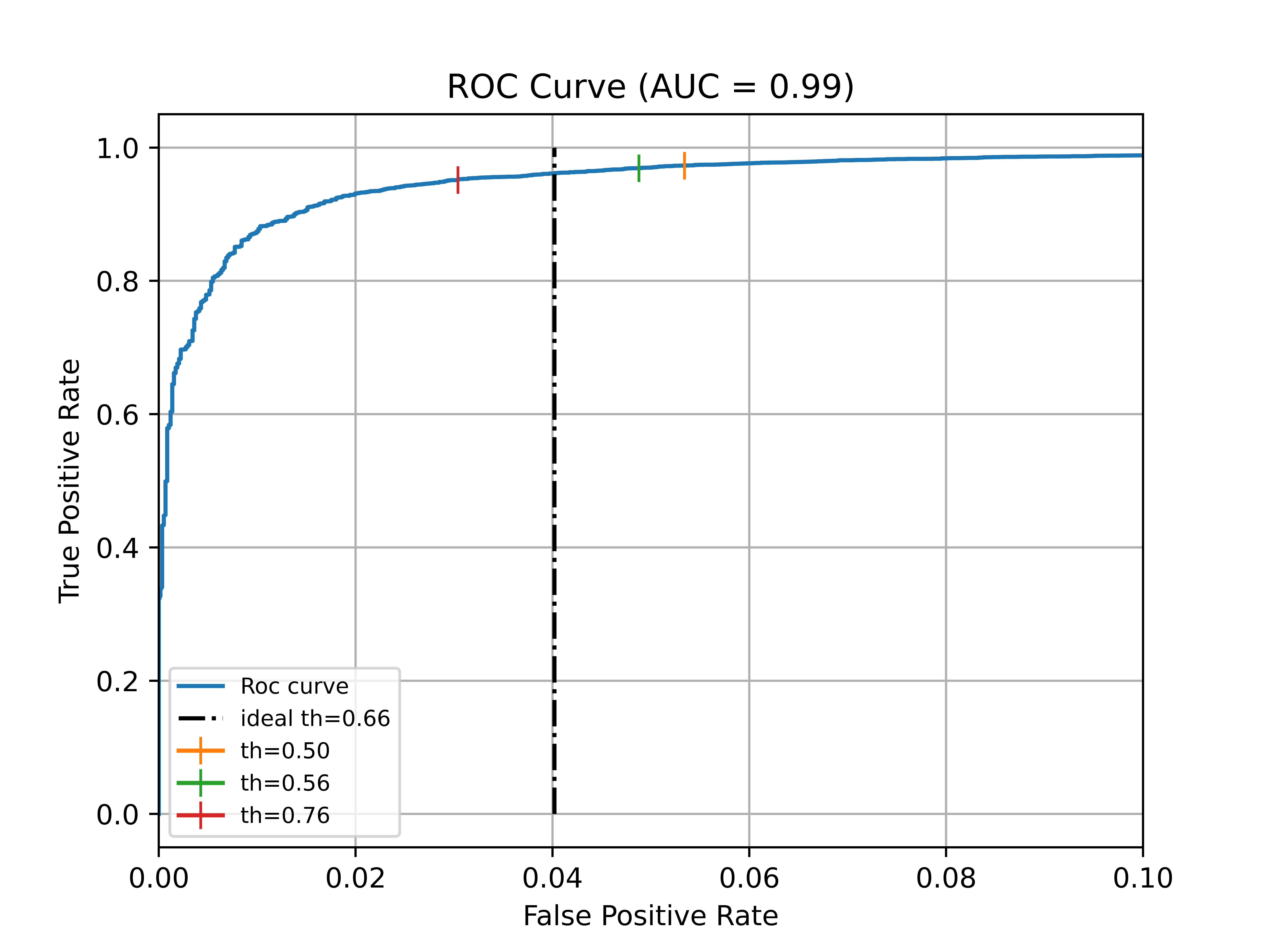}}
\caption{Zoomed ROC curve with threshold markings. \textbf{Orange:} default threshold = 0.5; \textbf{Black Dashed:} ideal threshold = 0.6587; \textbf{Green:} ideal\_th - 0.1 = 0.56; \textbf{Red:} ideal\_th + 0.1 = 0.76}
\label{fig:roc_curve_zoomed}
\end{figure}

The blue graph in Fig. \ref{fig:roc_curve} and \ref{fig:roc_curve_zoomed} shows the ROC curve for our model. The yellow dashed line indicating a random classifier serves as reference. On the ROC curve there are different thresholds marked by small ticks. First of all, the orange tick shows the default $th = 0.5$ and second, the ideal threshold $ideal\_th = 0.6587$ is marked by the red tick. In order to get a feeling on how fast the threshold changes along the ROC curve, two additional ticks (green and red) are plotted to mark the $ideal\_th \pm 0.1$. Nevertheless, the most important point on the ROC curve is the $ideal\_th$ which reflects that point on the curve that is closest to the top left corner. This is because a good classifier has a high $tpr$ and a low $fpr$. Thus, if the ROC curve would hit the top left corner, we would have built a perfect classifier which could classify 100$\,$\% of the input images without any error.

Another important value in the ROC space is the Area Under Curve (AUC). This parameter is simply the area between the ROC curve and the x-axis. The closer the ROC curve approaches the top left corner, the larger is the area under the ROC curve, the higher is the value for the $AUC$. Thus for a perfect classifier, the value would be $AUC=1$ \cite{key:Fawcett_2006}. Our model reaches an $AUC = 0.9931$. 

\subsection{Confusion Matrix}

\begin{figure}[htbp]
\centerline{\includegraphics[width=8cm]{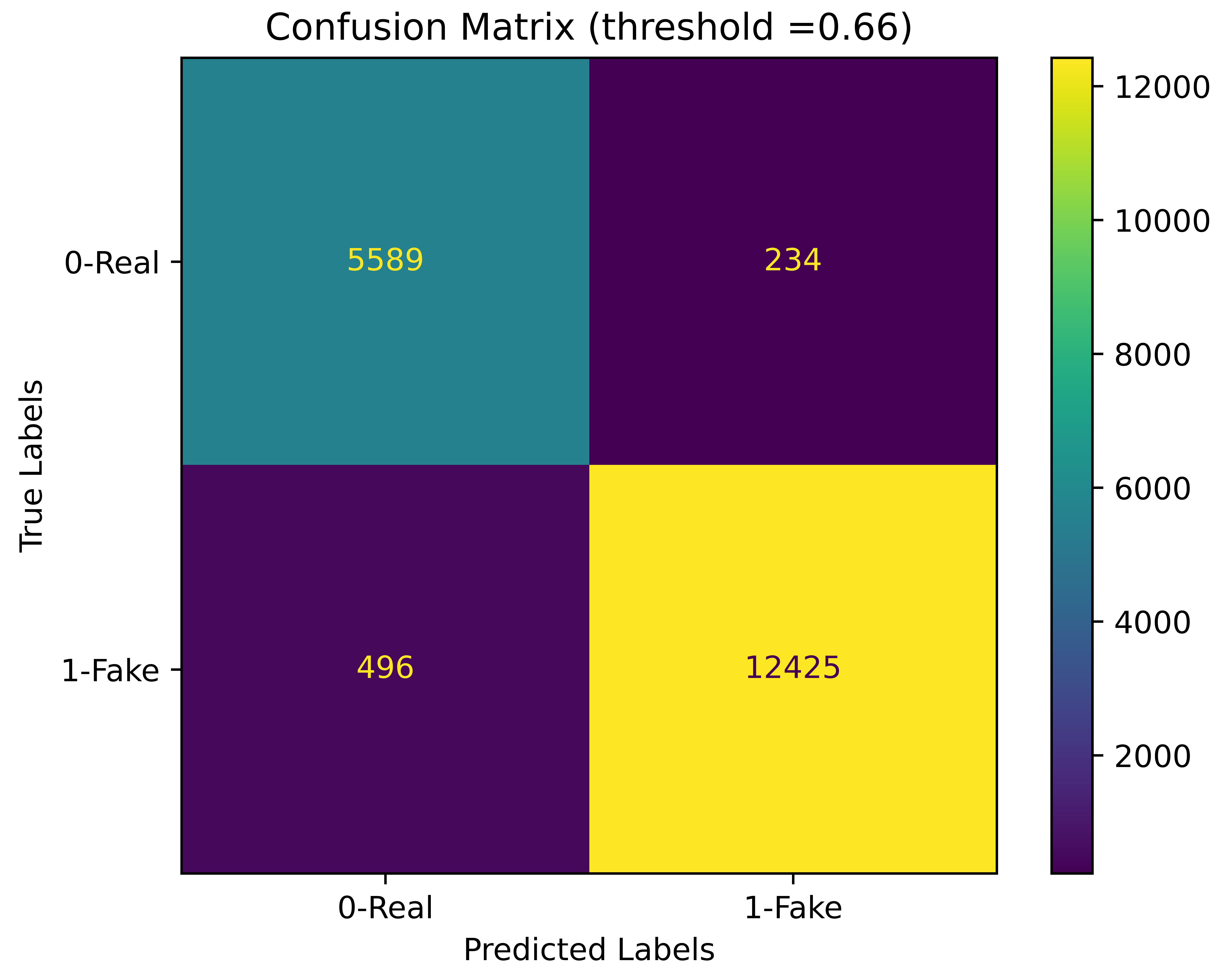}}
\caption{Confusion Matrix using ideal threshold}
\label{fig:cm}
\end{figure}

In order to determine the ideal threshold using the ROC curve, the number for $tp$, $fp$, $tn$, and $fn$ of the predicted images had to be calculated for every threshold iterated. The values on which the ideal threshold is based on, can be displayed in a so-called Confusion Matrix (shown in Fig. \ref{fig:cm}) and are used as a metric to evaluate the networks performance as well. On the x-axis, the predicted label $\hat{y}$ is shown, while on the y-axis, the actual label $y$ is displayed. For correctly predicted images, we reached a true positive value of $tp = 12,425$, for the true negative parameter a value of $tn = 5,589$. On the wrongly predicted images, the number of false positive predicted images reaches $fp = 234$ and the number of false negative predictions counts $fn = 496$.

\subsection{F1-Score}

Beside the already mentioned metrics for assessing the networks performance, the ROC curve, AUC parameter, and the Confusion Matrix, there are further metrics which can be calculated using the already available parameters $tp$, $fp$, $tn$, and $fn$. First of all, the accuracy referred to in equation \ref{eqn:acc} indicates the number of correctly predicted images divided by the total number of test images \cite{key:Fawcett_2006}. This metric is only reliable if the dataset used is balanced across both classes, the negative and the positive \cite{key:Longunova_2023}. Our test data set has, according to table \ref{tab:datasets_classes}, more than twice as many data points in the positive class as in the negative class. Thus it is considered as imbalanced which is the reason, why there are further metrics to be considered when evaluating the performance. Nevertheless, the accuracy value was calculated for our model to get a first impression on the performance, see table \ref{tab:results}.

\begin{equation}
\label{eqn:acc}
\text{accuracy} = \frac{tp + tn}{(tp + fn) + (tn + fp)} = \frac{tp + tn}{P + N}
\end{equation}

To indicate better, how precisely the model can predict the positive class, the precision metric (equation \ref{eqn:pre}) is used. It shows the proportion of correctly predicted positives by dividing the true positive value by all positive predictions. On the other hand, the metric recall (equation \ref{eqn:rec}), also referred to as $true ~positive ~rate$ as already shown in equation \ref{eqn:tpr}, divides the true positives by the number of images in the positive class. Thus it indicates the portion of the positive class which was correctly predicted. These two metrics, precision and recall, are competitive. Thus improving one of them means downgrading the other at the same time. \cite{key:Longunova_2023}

\begin{equation}
\label{eqn:pre}
\text{precision} = \frac{tp}{tp + fp}
\end{equation}

\begin{equation}
\label{eqn:rec}
\text{recall} = tpr = \frac{tp}{tp + fn} =\frac{tp}{P}
\end{equation}

Since these both metrics are competitive, maximize a 'combined' metric of both leads to an overall improved model. The F1-Score metric is calculated by using both, the precision and the recall, according to equation \ref{eqn:f_one}. The advantage over the accuracy is, that the F1-score considers the performance on each class individually while the accuracy only takes the positive class into account. The value for the F1-score results in a range from 0 to 1, whereas the higher the score is the better is the overall performance of the model \cite{key:Longunova_2023}.

\begin{equation}
\label{eqn:f_one}
\text{F1-score} = 2 \times \frac{\text{precision} \times \text{recall}}{\text{precision} + \text{recall}}
\end{equation}

%
%
%
%

In our case, we reached the following results as shown in table \ref{tab:results}.

\begin{table}[htbp]
\begin{center}
\caption{Performance Measures}
\label{tab:results}
\begin{tabular}{l|c|c|c}
Metric    & ideal\_th - 0.1 & Ideal Threshold & ideal\_th + 0.1 \\ \hline
Accuracy  & 0.9636 & 0.9611 & 0.9572 \\
Precision & 0.9779 & 0.9815 & 0.9858 \\
Recall    & 0.9690 & 0.9616 & 0.9516 \\
F1-Score  & 0.9734 & 0.9715 & 0.9684 \\
\end{tabular}%

\end{center}
\end{table}

\subsection{Probability Density Function}

\begin{figure}[htbp]
\centerline{\includegraphics[width=8cm]{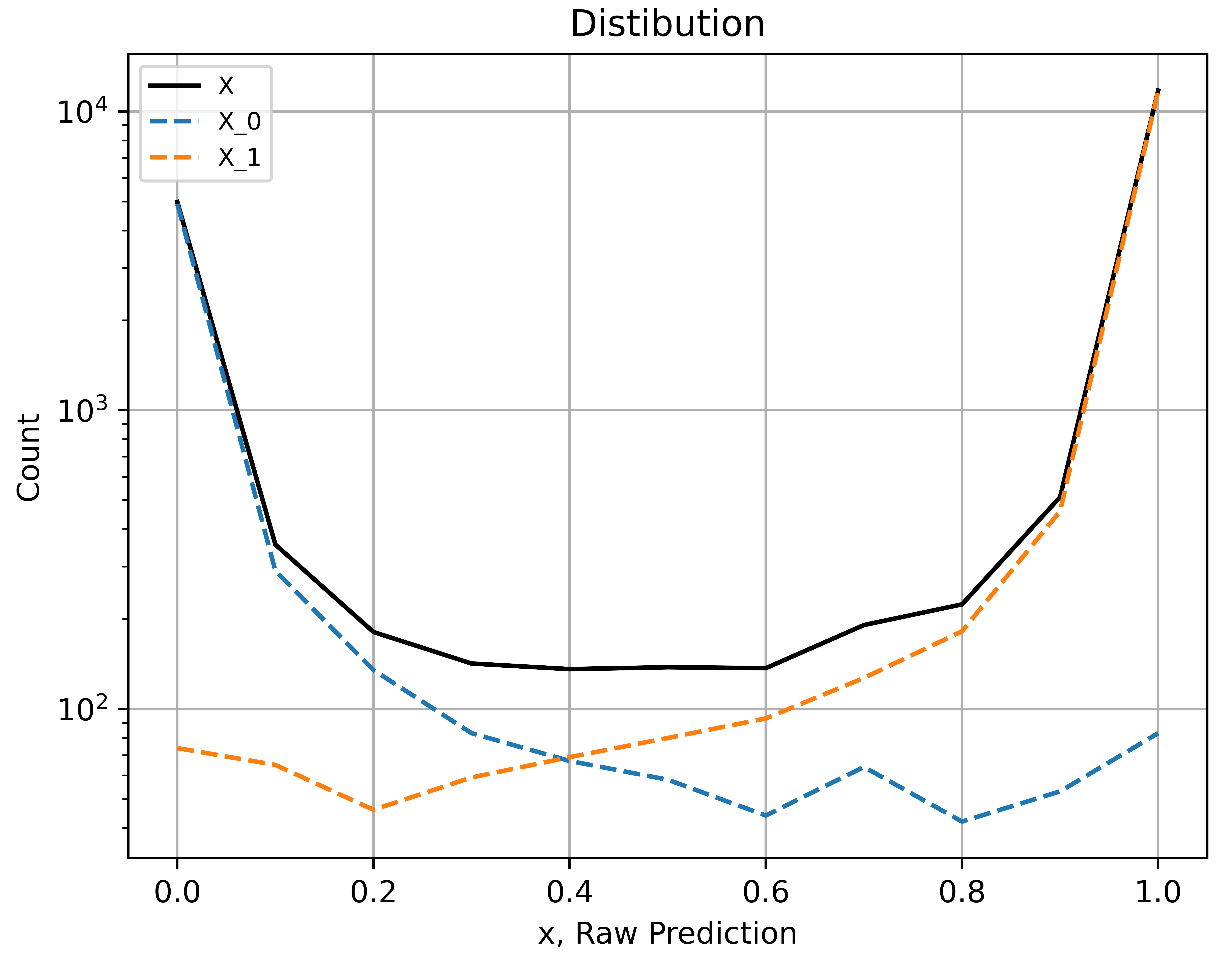}}
\caption{Distribution of Continuous Random Variable: \textbf{Black:} $X$. \textbf{Blue Dashed:} $X_0$. \textbf{Orange Dashed:} $X_1$}
\label{fig:count}
\end{figure}

All the metrics shown above assess the overall performance of our neural network. In order to focus more on how the model performs on each of the two classes, we want to have a look on the probability density function of the predictions. To do so, the outcome of the model, i.e., the prediction $\hat{y}_{raw}$, is considered as a continuous random variable $X$ which indicates the probability that the input image belongs to the positive class. As already shown, it applies $X = \hat{y}_{raw} \in [0,1]$. In order to have a closer look on each class, $X$ is split into $X_0$ and $X_1$. While the predictions done by the model for all images contained in the test data set are counted in $X$, the variables $X_0$ and $X_1$ contain only the prediction for those images with an actual label $y$ of the negative respective positive class. Thus it applies $X_0=\hat{y}_{raw}|y=0$ and $X_1=\hat{y}_{raw}|y=1$. Fig. \ref{fig:count} shows the distribution of all three variables by counting the occurrences of predictions along x as raw prediction. Since X was split up, it applies that $X = X_0 + X_1$

Based on these continuous random variables, the Probability Distribution Function (PDF) is calculated. The real PDF can not be determined since we have only discrete predictions as an output from the model. Therefore Kernel Density Estimation (KDE) based on the predictions as samples is used to get an estimation close to the real PDF \cite{key:Hastie_2009}. Let $f(x)$ be the real PDF, $\hat{f}(x)$ is the KDE.

We use the function 'kdeplot' from seaborn library to plot the KDE. It is calculated according to equation \ref{eqn:kde} \cite{key:Chen_2017}. This equation calculates the probability density for one specific point x. Here, $n$ is the number of data points (images in test data set: $n=18,744$), $h$ is the bandwidth which serves as a smoothing parameter, and $K$ is the kernel function. In this case, the Gaussian Kernel according to equation \ref{eqn:gauss_kernel} \cite{key:Chung_2007} is applied. For the smoothing parameter, kdeplot is not limited to a specific value. The Seaborn library uses an adaptive bandwidth parameter $h$ based on the Scott method \cite{key:seaborn_2024}.

\begin{equation}
\label{eqn:kde} 
\hat{f}(x) = \frac{1}{n h} \sum_{i=1}^{n} K\left( \frac{x - x_i}{h} \right)
\end{equation}

\begin{equation}
\label{eqn:gauss_kernel} 
K(x) = \frac{1}{h\sqrt{2\pi}} \mathrm{e}^{\left(-\frac{x^2}{2h^2}\right)}
\end{equation}

Since the focus is on each individual class, also the probability distribution function is split, like done for the variable X. Thus $\hat{f_0}(x)$ shows the KDE for test images which belong to the negative class, whereas $\hat{f_1}(x)$ is the complementary part for the positive class. See equations \ref{eqn:pdf_neg} and \ref{eqn:pdf_pos} for detail.

\begin{equation}
\label{eqn:pdf_neg} 
\hat{f_0}(x) = \hat{f}(x|y=0)
\end{equation}

\begin{equation}
\label{eqn:pdf_pos} 
\hat{f_1}(x) = \hat{f}(x|y=1)
\end{equation}

\begin{figure}[htbp]
\centerline{\includegraphics[width=8cm]{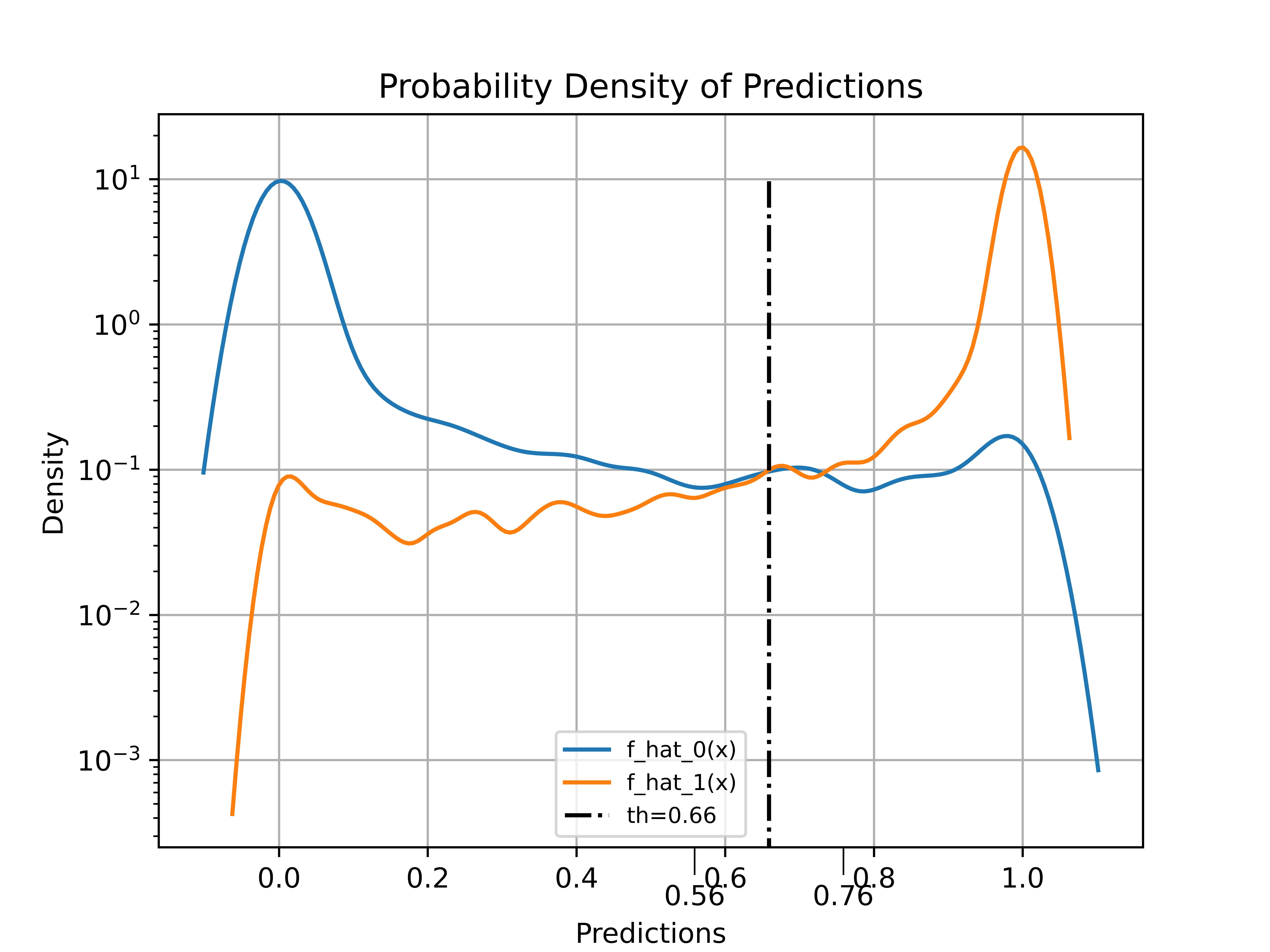}}
\caption{Kernel Density Estimation of Predictions for Positive Class $\hat{f}_1$ and Negative Class $\hat{f}_0$}
\label{fig:pdf}
\end{figure}

Thus the KDE plots for $\hat{f_0}(x)$ and $\hat{f_1}(x)$ shown in Fig. \ref{fig:pdf}, display the probability density for variables $X_0$ and $X_1$ over the predictions output of the model (on the x-axis) which indicates the probability that the input belongs to the positive class. Since we display the probability distribution for both classes separately, we see on the one hand, how likely it is that the model predicts a high probability given that the actual class of the input image is positive. On the other hand, we see how likely it is that the model predicts a low probability give that the actual class is negative. 

In addition, the ideal threshold is shown as dashed line, and also $ideal\_th \pm 0.1$ are marked as small ticks on the x-axis in alignment to the ROC curve shown in Fig. \ref{fig:roc_curve}. Based on the chosen ideal threshold, every predictions higher than the threshold is counted as predicted positive class while everything below is counted as predicted negative class. Therefore combining the KDE plots with the ideal threshold, it can be seen, how likely it is that the model predicts a positive or negative class. Since the KDE plots are separated for both classes, the figure displays how likely it is that the model predicts either the correct or the wrong class in combination with the actual class.

\section{Discussion}
\subsection{Discussion of Results}
\paragraph{ROC curve}

The ROC curve of our model in Fig. \ref{fig:roc_curve} shows a nearly error-free classifier. A perfect classifier would be indicated by an AUC value of 1 and would hit the top left corner of the plot. We reached a AUC value of 0.9931 with our model, nearly reaching the top left corner. This shows the thorough capability of our model regarding classifying the test images as authentic or fake. 

Furthermore, our model performs stable with respect to the chosen threshold. Fig. \ref{fig:roc_curve} marks four different threshold values ($th \in \{0.5, 0.56, 0.66, 0.76\}$) as an example. These four markings are very close to each other. This means, even if we choose another value for the threshold, different to the ideal one calculated by the ROC curve, the performance of the network in terms of confusion matrix and F1-score would only be slightly affected. This observation is based on the fact, that our model is confident in its predictions, even if wrongly predicted, meaning most of the predictions are close to 0 or 1 while the absolute minority of the predictions is in between (also displayed in Fig. \ref{fig:count}).

\paragraph{Confusion Matrix}
The confusion matrix in Fig. \ref{fig:cm} shows the values for correct and wrong predictions for the negative and the positive class. It can be seen that our model predicts only the absolute minority, in detail 3.89$\,$\%, of all images wrongly. This reflects that the model does not only focus on predicting one class correctly while ignoring the other one, but that the model predictions are stable over both classes.

A second remarkable aspect here is, that the number for $false ~positive$ predicted images is less than half of the number for $false ~negative$ predictions. On the one hand this can be explained due to the number of images of the positive class (i.e., fake images) in the test data set, which is twice the number of images in the negative class. On the other hand this effect is based on the selection of the ideal threshold which focuses on the $false~positive~rate$ and tries to minimize it.

\paragraph{Probability Density Function}
Plotting the PDF separated for the positive and the negative class as in Fig. \ref{fig:pdf}, shows the following four interesting effects.

First, we would have expected that the PDF plot for the corresponding class only shows an increase of the value at the location the correct label. This means that the plot for the negative class ($\hat{f}_0$) increases its value only at x=0 and the plot for the positive class ($\hat{f}_1$) only at x=1. Nevertheless, an increase of the value at the opposite class label for both plots can be recognized. This might be due to the sigmoid activation function used for the output neuron together with the training process that uses discrete labels (i.e., 0 or 1) to tell the network what would have been the correct output.

Second, the first intersection between the two plots is surprisingly close to, if not even with, the location of the ideal threshold. This might be because of the calculation of the ideal threshold. It is determined using the ROC curve by maximizing the $true ~positive ~rate$ and minimizing the $false ~positive ~rate$. The intersection of the two plots probably marks the best trade off between these two values and is therefore calculated as ideal threshold.

Third, the effect that the PDF is also plotted for values below 0 and above 1 even though the model prediction is limited to [0,1] is caused by the Gaussian Kernel which is used to calculate the KDE values.

Fourth, the fact that the peak values for the positive class are higher than for the negative class can be explained with the imbalanced test data set, which includes more than twice as many data points for the positive class than for the negative.

\subsection{Discussion about the Model}
\paragraph{Building the Model}
We built our model out of two parts: first a base model and second, own layers attached to the base model which are adopted to the task of binary classification. As base model, the ResNet-50 is chosen, since it is very effective in image classification task due to the following reasons: it addresses the problem of vanishing gradients, by the use of residual blocks and especially the contained skip connections which make the training of deep neural networks more efficient. Moreover these blocks consist of convolutional layers which are designed for image processing \cite{key:Ibrahim_2024} \cite{key:Garg_2022}.

By replacing the last 1000-d fully connected layer from ResNet-50 with an own dense layer, we successfully followed the approach shown by Rezende et al. \cite{key:Rezende_2017} and by Reddy and Juliet \cite{key:Reddy_2019} for binary classification tasks.

\paragraph{Training}
For the training process, transfer learning using pre-trained parameters of the ResNet-50 is applied because the number of images in the collected data sets is too small for a whole network to be trained completely with randomly initialised parameters. Instead, the parameters are pre-trained on the ImageNet dataset. Transfer learning is a commonly used process in machine learning, also used together with the ResNet model for example in \cite{key:Rezende_2017} and \cite{key:Reddy_2019}. 

Usually, transfer learning is applied between two tasks that share low-level abstraction of visual shapes like edges but have different output requirements. In such cases, the upper layer in the neural net are locked meaning the parameters are not trainable but only the lower layers including the very last layer are trainable and thus updated and adopted to the new task during training \cite{key:Bengio_2016}. However, we reached the best results when all layers are trainable in the first training step. Re-training a whole net for a different data set is called fine tuning. Yosinski and Clune showed in \cite{key:Yosinski_2014} that transferring features (i.e., using pre-trained parameters) together with fine tuning the whole net boosts the generalization performance of the network. According to their research, keeping more pre-trained layers and fine tune them afterwards, improves the performance slightly better than keeping less layers. This effect even applies for large target data sets \cite{key:Bengio_2016}. Based on this, we think that fine tuning the ResNet-50 for our task also improves the performance of the neural network in total.

As explained in section \ref{sec:methods}, after the first step of fine tuning the net, two further steps followed with less layers set to trainable. We did this in order to apply 'normal' transfer learning to our network and to train our own added dense layer at the output of the net even further.

Looking at Fig. \ref{fig:train_one} (training step 1), the validation loss was reduced in general during training until the implemented early stopping functionality stopped the training 5 epochs later. Nevertheless, based on the validation loss, the training process is a bit unstable which is shown as high values for the validation loss for some of the epochs. It might be that the data set used is too small for training the whole ResNet-50 network. For step 2 (Fig. \ref{fig:train_two}) the validation loss was only improved for one epoch and started increasing afterwards while for step 3 (Fig. \ref{fig:train_three}) it was not improved at all. This effect was not investigated further but it seems that after fine tuning a whole neural network applying transfer learning is superfluous and does not improve the networks performance. Since step 2 and 3 do not significantly improve the validation loss, step 1 would have been enough to carry out.
\section{Conclusion}

We built a simple neural network model for classifying human portraits as authentic or fake. Based on the ResNet-50 model which contains convolutional layers designed for image processing, we added our own layers to adopt it to our task. The single output neuron indicates a probability of the authenticity of the input image. For the transfer learning, we used the 'Diverse Face Fake Dataset' containing a wide range of different manipulation methods and also diversity in terms of faces visible on the images. With our final model we are able to correctly classify over 96$\,$\% of our test data and reach the following remarkable performance metrics: precision = 0.98, recall 0.96, F1-Score = 0.97 and an AUC of 0.99.

We saw that, when using fine tuning for training the whole model at once, subsequent transfer learning only adds a minor improvement to the models performance in terms of minimizing the validation loss. Thus the second and the third training step were nearly superfluous and could have been skipped at all. Nevertheless, these additional training steps had no negative influence on our model.

Future researches may have a more detailed look on the probability distribution function (or in particular the kernel density estimation) in combination with the ideal threshold calculated using the ROC curve. When plotting two separated graphs for the KDEs of the two classes (positive and negative), at a certain point an intersection occurs. As shown, it seems to be the case, that this intersection is close to, if not even with the ideal threshold by the ROC curve. This intersection could be the perfect trade-off between the true positive rate and the false positive rate which is also calculated using the ROC curve. This hypothesis was not analysed in detail but might be subject of future researches. 

In addition, an interpretation of the model could be interesting and revealing as well. Researches could focus on the question, why does the model predict a given input image as authentic or fake. In detail, on which features of the portrait is this prediction based on and which areas of an image have the most impact on this prediction?

\section{Acknowledgment}
We would like to express our thanks to our lab engineer Mr. Gerald Schickhuber, for his technical assistance in setting up the used PC, without which this project would not have been possible.

\renewcommand{\refname}{References}

\bibliographystyle{IEEEtran}
\bibliography{tex/Bibliographie}

\appendix
\subsection{Matlab Alignment}

\begin{figure}[htbp]
\centerline{\includegraphics[width=8cm]{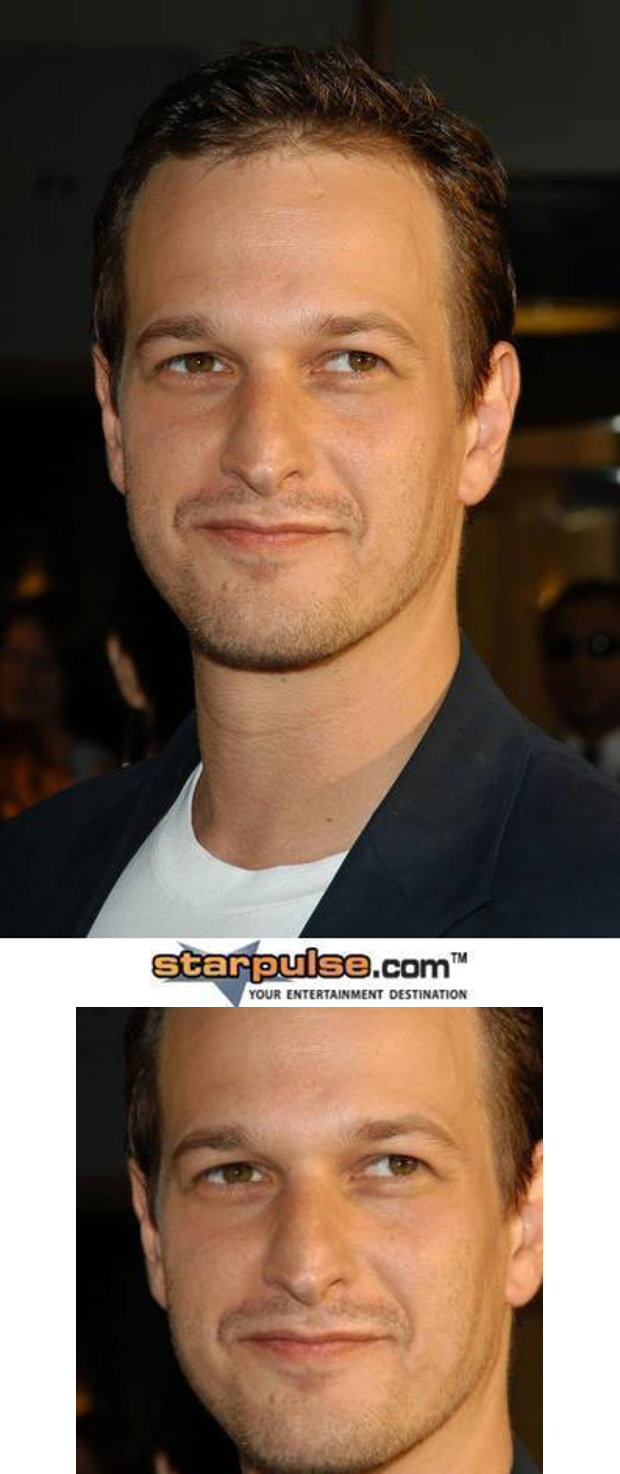}}
\caption{Sample Image for Matlab Alignment. \textbf{Top:} Before; \textbf{Bottom:} After \cite{key:Liu_2015}}
\label{fig:alignment}
\end{figure}

\newpage
\subsection{Deselected Images}

\begin{figure}[htbp]
\centerline{\includegraphics[width=7cm]{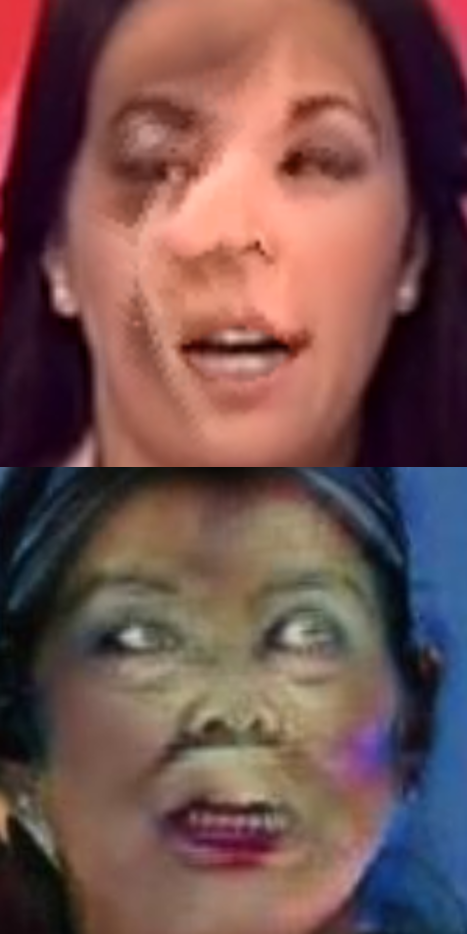}}
\caption{Unused Images. \textbf{Left:} Faceswap \cite{key:Rossler_2019}; \textbf{Right:} StarGAN \cite{key:yunjey_2018}}
\label{fig:unused}
\end{figure}

\begin{figure*}[htbp]
\subsection{Datasets}
\centerline{\includegraphics[width=18cm]{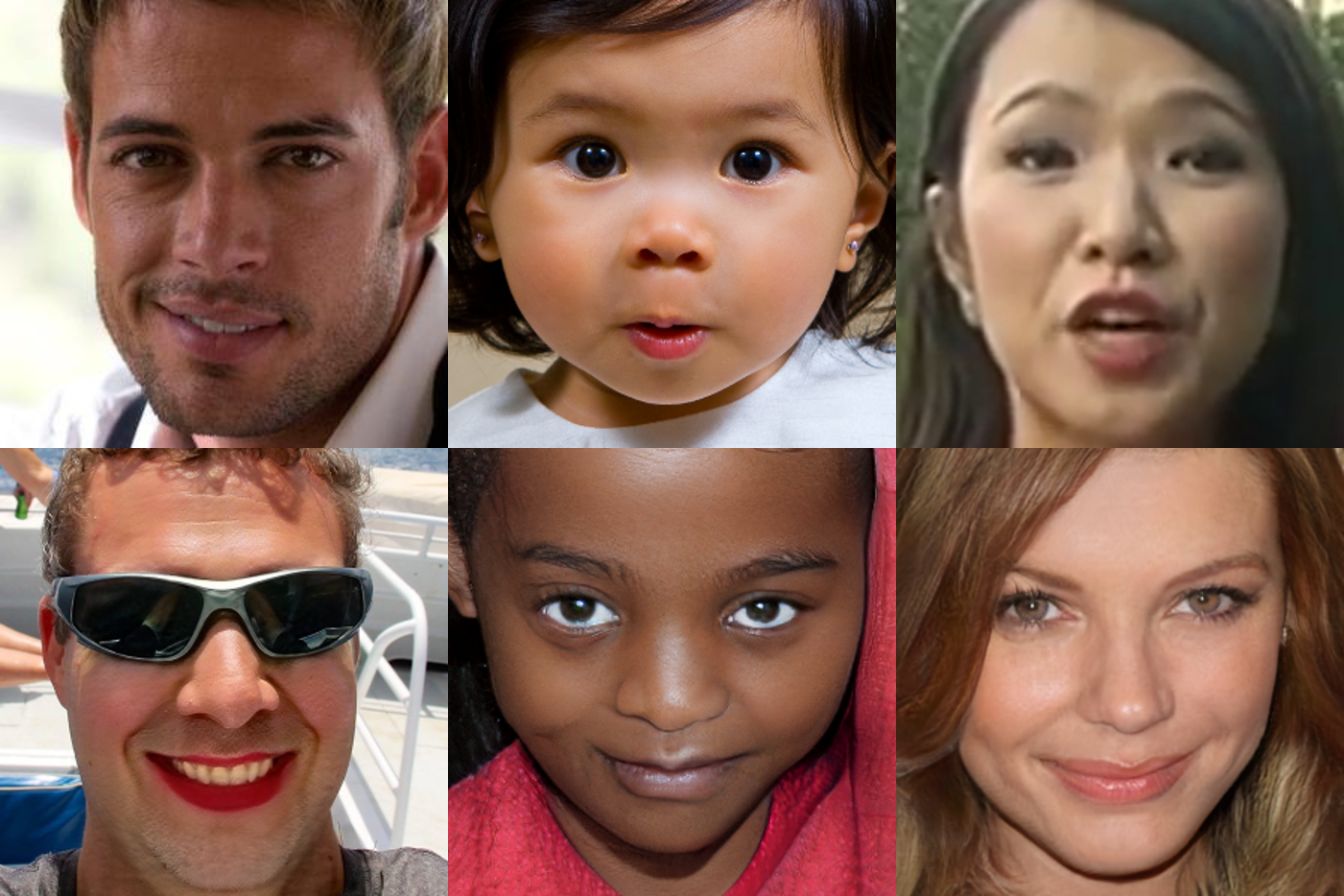}}
\caption{Samples for Different Datasets. \textbf{Top Row - Real:} CelebA \cite{key:Liu_2015}, FFHQ \cite{key:karras_2018}, FaceForensics++ \cite{key:Rossler_2019}; \textbf{Bottom Row - Fake:} FaceApp \cite{key:faceapp_2019}, Style GAN FFHQ \cite{key:karras_2018}, PG GAN \cite{key:karras_2017}}
\label{fig:datasets}
\end{figure*}

\end{document}